\documentclass[sigconf]{acmart}

\usepackage{subfig}
\usepackage{hyperref}

\copyrightyear{2021}
\acmYear{2021}
\setcopyright{acmlicensed}\acmConference[GECCO '21]{2021 Genetic and Evolutionary Computation Conference}{July 10--14, 2021}{Lille, France}
\acmBooktitle{2021 Genetic and Evolutionary Computation Conference (GECCO '21), July 10--14, 2021, Lille, France}
\acmPrice{15.00}
\acmDOI{10.1145/3449639.3459321}
\acmISBN{978-1-4503-8350-9/21/07}

\begin{document}

\title{Monte Carlo Elites: Quality-Diversity Selection as a Multi-Armed Bandit Problem}

\author{Konstantinos Sfikas}
\affiliation{%
  \institution{University of Malta}
  \city{Msida}
  \country{Malta}}
\email{konstantinos.sfikas@um.edu.mt}

\author{Antonios Liapis}
\affiliation{%
  \institution{University of Malta}
  \city{Msida}
  \country{Malta}}
\email{antonios.liapis@um.edu.mt}

\author{Georgios N. Yannakakis}
\affiliation{%
  \institution{University of Malta}
  \city{Msida}
  \country{Malta}}
\email{georgios.yannakakis@um.edu.mt}

\begin{abstract}
A core challenge of evolutionary search is the need to balance between exploration of the search space and exploitation of highly fit regions. Quality-diversity search has explicitly walked this tightrope between a population's diversity and its quality. This paper extends a popular quality-diversity search algorithm, MAP-Elites, by treating the selection of parents as a multi-armed bandit problem. Using variations of the upper-confidence bound to select parents from under-explored but potentially rewarding areas of the search space can accelerate the discovery of new regions as well as improve its archive's total quality. The paper tests an indirect measure of quality for parent selection: the survival rate of a parent's offspring. Results show that maintaining a balance between exploration and exploitation leads to the most diverse and high-quality set of solutions in three different testbeds.
\end{abstract}



\maketitle

\section{Introduction}\label{sec:introduction}

Population-based genetic algorithms are powerful when it comes to global optimization, but have often faced the challenge of premature convergence towards local optima in the search space \cite{goldberg1987deceptive,naudts1999epistasis}. Many approaches across several decades have attempted to address this challenge, including genetic diversity preservation mechanisms \cite{martin2000island,stanley2002neat}, multimodal optimization \cite{preuss2015multimodal}, multi-objective approaches \cite{dejong2001reducing} and many others. For over a decade, divergent search has also been a prominent solution to premature convergence by ignoring the fitness function and instead aiming to maximize the population's behavioral diversity \cite{lehman2011abandoning}. Divergent search favors exploration of the search space exclusively, and has shown to work particularly well in deceptive fitness landscapes \cite{liepins1991deceptiveness}. Somewhere between the greedy exploitation strategy of objective-driven methods and the exploration-based divergent search lies the Quality-Diversity (QD) family of algorithms \cite{pugh2016quality,cully2018qualitydiversity} which attempt to balance exploration and exploitation by guiding evolution towards behavioral diversity while also rewarding individuals that are better (in terms of the objective) in their own behavioral niches. This localized control for quality is facilitated by ensuring that individuals satisfy some minimal constraints \cite{lehman2010revising,liapis2015ecj}, by treating diversity and local competition as separate objectives \cite{lehman2011creatures}, or by partitioning the search space in advance and saving only the best individual per partition \cite{Mouret2015IlluminatingSS}.

As noted above, evolutionary computation has a long history in exploring the trade-offs between exploration and exploitation. Among QD approaches specifically, novelty search with local competition \cite{lehman2011creatures} explicitly treats a measure of exploration and a measure of exploitation as different objectives. On the other hand, the Multi-dimensional Archive of Phenotypic Elites (MAP-Elites) algorithm ensures exploration through the way the search space is partitioned; in this vein, alternative ways of partitioning the space have been explored via clustering \cite{vassiliades2017voronoi,fontaine2019sliding} or dimensionality reduction \cite{cully2019autonomous}. Most implementations of MAP-Elites select individuals uniformly among the elites retained in the feature map, and a pressure towards exploitation comes during replacement rather than during parent selection. While MAP-Elites keeps only the fittest individual in each cell, variants with more individuals per cell have been explored \cite{khalifa2018talakat, flageat2020deep, justesen2019adaptive}. Yet parent selection is an important way to ensure exploitation that has only been investigated in few studies---e.g. prioritizing parents that are novel \cite{pugh2016quality} and/or surprising \cite{gravina2019blending} in MAP-Elites. Cully and Demiris explored the impact of parent selection on QD search in an extensive study \cite{cully2018qualitydiversity}.

This paper investigates the traditional exploration-exploitation dilemma in QD by transferring well-studied node selection approaches from tree search \cite{kocsis2006bandit} to parent selection strategies in quality diversity. In particular, in this study we view the selection process of MAP-Elites as a multi-armed bandit problem and we explore how the upper confidence bound (UCB) formula can drive parent selection, specifically the variant popularized in Monte Carlo Tree Search \cite{kocsis2006bandit} (i.e.~UCB1 applied to trees). Since the UCB formula explicitly designates an exploitation and an exploration component, we also test how these two components impact the performance of MAP-Elites on their own. Importantly, the exploitation component is not directly fitness-dependent but instead rewards individuals that produce offspring that survive, similar to the curiosity score of \cite{cully2018qualitydiversity}. Finally, we introduce two UCB score variants, where the formula is calculated on the parent as an individual or on the cell which is occupied by the parent. The many variants of this selection mechanism are evaluated comprehensively in two popular testbeds for QD search and on a simple level generation task. Results on the two testbeds show that all variants outperform the ``vanilla'' uniform selection strategy of the algorithm in terms of all established metrics for QD search. Moreover, UCB is shown in all cases to better balance highly fit and diverse solutions versus e.g. many diverse solutions found by exploitation-only methods.

\section{Background}\label{sec:background}

The MAP-Elites algorithm was introduced by Mouret and Clune \cite{Mouret2015IlluminatingSS} as a way to illuminate the search space during optimization. MAP-Elites maintains a diverse set of high-quality individuals and is one of the prototypical QD search methods. MAP-Elites uses a feature map based on $D$ feature dimensions that describe the phenotype or its behavior. This feature map is partitioned into a number of cells depending on the resolution chosen, and each cell may contain up to one individual. Each individual on the feature map is thus guaranteed to have at least one behavioral characteristic different from every other individual. Each individual is evaluated across the $D$ feature dimensions, and on a problem-specific fitness function (the quality component of the algorithm). If the individual is mapped to an unoccupied cell of the feature map, it occupies it; if it is mapped to a cell already occupied by an individual, it replaces that individual if its fitness is better (i.e. higher for maximization problems). This ensures that the feature map contains elites in each behavioral niche. As a summary of each iteration in MAP-Elites: (a) the algorithm randomly selects an individual among those in the feature map, (b) the individual produces an offspring through mutation, (c) the offspring is evaluated across all feature dimensions and fitness, (d) the offspring is mapped to a cell of the feature map and occupies it if the cell is empty or if the individual in that cell (which could be its parent) has a worse fitness.

As noted in the introduction, the general principle of MAP-Elites (partitioning the space and storing the fittest individual in each partition) has inspired a large number of variants. In terms of parent selection pressure, early work explored how a novelty score could be used to apply selection pressure based on the average behavioral distance of the individual and its nearest neighbors in an archive of past solutions \cite{pugh2016quality}. This was expanded in \cite{gravina2019blending} which biased parent selection according to a novelty score, a surprise score based on deviations from predicted trends in the population, and aggregated or multi-objective combinations of the two. Cully and Demiris \cite{cully2018qualitydiversity} explored several selection methods, including score-based and population-based (where multiple offspring are inserted simultaneously to the feature map). Criteria for selecting parents included their fitness, their novelty (which, unlike \cite{lehman2011abandoning,pugh2016quality,gravina2019blending}, was defined as the number of filled cells neighboring the individual) and a curiosity score. The latter rewarded individuals based on the number of offspring that survived and penalized them based on the number of offspring that did not. All of these scores were applied, individually, in a score-proportionate, stochastic selection process; a multi-objective variant was also tested with fitness and novelty as separate objectives. All of these metrics could in principle be used as the exploitation dimension for a UCB-based selection method. The curiosity score in particular is very similar to the offspring survival metric used here (essentially with 0 penalties for offspring that perish), although there are two important differences in terms of the broader selection process: (a) the selection bias formula used here normalizes the offspring to the number of times an individual is selected, thus applying immense pressure to newly added individuals (or cells) while the opposite is true with curiosity-based selection; (b) parent selection in this paper is performed by ranking individuals, rather than via stochastic selection (random selection is only used to break ties). Other variant parent selection mechanisms include cases where multiple feature maps are kept, such as two feature maps with different feature dimensions \cite{pugh2016quality} or containing feasible individuals in one and infeasible individuals in the other \cite{khalifa2018talakat}; in these cases, an equal amount of parents are chosen from each feature map. Finally, Go-Explore \cite{ecoffet2021goexplore} applies a selection bias aggregating different versions of selection frequency (including e.g. the number of times the cell is selected since its offspring discovered a new cell) as well as neighborhood- and domain-dependent biases. The selection bias of Go-Explore is fairly close in principle to the UCB formula presented here, as the latter also considers the survival rate of offspring (although survival also considers an offspring replacing an existing elite rather than only discovering new cells). Similar to UCB, the Go-Explore selection formula is also a weighted sum of components, some of which could be construed as `exploitation' and `exploration' measures. Notably, the proposed UCB formula is not dependent on the domain and is tested on very different problems. It also applies a strict rank-based selection priority rather than the probabilistic approach of Go-Explore.

To the best of our knowledge, the UCB formula has not been utilized for parent selection in MAP-Elites and its variations. The closest application of UCB policies to our work is in surrogate-assisted illumination (SAIL), which used the UCB formula to select which individuals should be simulated \cite{gaier2017sail}. The selection was performed on an acquisition map (i.e. a map containing the predictions of a surrogate model regarding the fitness of the individuals). The acquisition map was in essence very similar to a feature map in MAP-Elites, although in SAIL the feature map is produced through predictions of the model (which is trained via the acquisition map) and during evolution parent selection is actually uniform. Moreover, the UCB formula in \cite{gaier2017sail} takes into account performance and variance of performance, rather than traces of the evolutionary progress as in this paper. In other work \cite{gaier2020discovering}, UCB1 is applied to choose which of the genetic operators (re-constructive crossover, line mutation, or isometric mutation) to apply to the parent. Similar to this paper, the reward in \cite{gaier2020discovering} ``is assigned in proportion to the number of children who earned a place in the archive''.

\section{UCB for Parent Selection}\label{sec:methodology}

This paper explores how parent selection driven by the UCB formula  \cite{kocsis2006bandit} can affect the performance of MAP-Elites. To test the general applicability of our approach, we use the original MAP-Elites implementation of \cite{Mouret2015IlluminatingSS} and we modify only the selection strategy. Unless specifically stated, parent selection methods described here rank all elites in the feature map based on a selection score, and choose the individual with the highest score; in case of ties, selection is random among the tied individuals.

The UCB formula of Eq.~\eqref{eq:selection_ucb} is applied to calculate the selection score, where $n(i)$ is the times individual $i$ was selected, $w(i)$ is the number of offspring of $i$ that survive (i.e. replace an existing individual or occupy an empty cell), $\lambda$ is a constant, and $N_s$ is the total number of selections for the whole population. We treat the edge-case of $n(i) = 0$ as ``infinity'', following a common interpretation of UCB \cite{Browne2012SurveyOfMCTSMethods}. In doing so, the algorithm is forced to visit every individual at least once, by giving equal and absolute priority to unvisited ones. In experiments presented in this paper, $\lambda=\frac{1}{\sqrt{2}}$ for UCB selection strategies, as this value is optimal when the reward ($w(i)/n(i)$) is in the value range of $[0,1]$ , as explained in \cite{Browne2012SurveyOfMCTSMethods}.
\begin{equation}\label{eq:selection_ucb}
U(i) =
\begin{cases}
\dfrac{w(i)}{n(i)} + \lambda\cdot\sqrt{\dfrac{ln(N_s)}{n(i)}} &\text{ if } n(i) > 0,\\
\infty &\text{ if } n(i) = 0,\\
\end{cases}
\end{equation}

In Eq.~\eqref{eq:selection_ucb}, it is equally valid to consider the times the individual has been selected as parent or the times any individual in that cell has been selected. We treat the former as \emph{individual-based selection}, $U_i(i)$, where $n_i(i)$ is the number of selections of this individual and $w_i(i)$ the times that offspring of this individual survived. We treat the latter as \emph{cell-based selection}, $U_c(i)$, where $n_c(i)$ and $w_c(i)$ is the number of instances that any elite occupying this cell was selected and produced offspring that survived respectively. Individual-based selection assumes that some individuals have strong potential to produce either highly fit (replacing existing elites) or highly diverse individuals (occupying new cells). Cell-based selection assumes that there are some inherent properties of that region of the search space that must be exploited. Importantly, cell-based selection ensures that each cell is selected at least once but new individuals in already occupied cells are not guaranteed to be selected.

This paper also explores the impact of each individual component of the UCB formula. Focusing on the exploitation-only methods, the $E_i(i)$ and $E_c(i)$ selection metrics use the same formula as Eq.~\eqref{eq:selection_ucb} with $\lambda=0$. For the exploration-only methods, Eq.~\eqref{eq:selection_exploration} calculates the metrics as a simplification of the second component of Eq.~\eqref{eq:selection_ucb}. Once again, two variants are calculated, the individual-based $X_i(i)$, and the cell-based $X_c(i)$, by setting $n(i)$ as $n_i(i)$ and $n_c(i)$ respectively.
\begin{equation}\label{eq:selection_exploration}
X(i) =
\begin{cases}
\dfrac{1}{n(i)} &\text{ if } n(i) > 0,\\
\infty &\text{ if } n(i) = 0,\\
\end{cases}
\end{equation}

In sum, six new selection metrics are tested in this paper, using variants of the UCB formula and measuring exploitation as the survival rate of the offspring of individuals. Three individual-based metrics consider the number of selections and times the offspring survived based on the current elite occupying this cell, while three cell-based metrics consider the number of selections and times the offspring survived based on every elite that ever occupied this cell. 
Some of these metrics are similar to existing methods: $X_c$ is conceptually similar to uniform selection but puts an emphasis on newly filled cells in the feature map. Later in the evolutionary process when all cells are filled, this pressure is less pronounced. $U_i$ selects parents only based on the ratio of surviving offspring of a specific individual, which is conceptually similar to the curiosity score of \cite{cully2018qualitydiversity}. However, there are two important differences: (a) the formula of Eq.~\eqref{eq:selection_ucb} gives exclusive priority to new individuals (when $n(i)=0$) and (b) this implementation always selects the individual with the highest offspring survival rate, versus the curiosity score-proportionate roulette wheel selection of \cite{cully2018qualitydiversity}. These two differences can skew selection substantially as evidenced in this paper's results.

We test the new selection metrics against three baselines. The \textbf{greedy} baseline ranks all elites by fitness and selects the fittest one to produce offspring. The \textbf{uniform} selection selects randomly any elite as in the original implementation of MAP-Elites \cite{Mouret2015IlluminatingSS}. The \textbf{curiosity} baseline follows the implementation of \cite{cully2018qualitydiversity}. Unlike all other methods (except uniform), the curiosity baseline uses roulette wheel selection proportionate to a curiosity score that increases by 1 when an individual's offspring survives and decreasing by 0.5 when it does not. In summary, the list of selection metrics examined is as follows:
\begin{itemize}
    \item $U_i$: individual-based UCB, with $\lambda=1/\sqrt{2}$ in Eq.~\eqref{eq:selection_ucb}.
    \item $U_c$: cell-based UCB, with $\lambda=1/\sqrt{2}$ in Eq.~\eqref{eq:selection_ucb}.
    \item $E_i$: individual-based exploitation-only, with $\lambda=0$ in Eq.~\eqref{eq:selection_ucb}.
    \item $E_c$: cell-based exploitation-only, with $\lambda=0$ in Eq.~\eqref{eq:selection_ucb}.
    \item $X_i$: individual-based exploration-only via Eq.~\eqref{eq:selection_exploration}.
    \item $X_c$: cell-based exploration-only via Eq.~\eqref{eq:selection_exploration}.
    \item $G$: greedy baseline selecting the fittest elite.
    \item $R$: ``vanilla'' MAP-Elites applying uniform (random) selection.
    \item $C$: roulette-wheel selection proportionate to the curiosity score formula of \cite{cully2018qualitydiversity}. As in the original paper, curiosity score is calculated per individual (not per cell).
\end{itemize}

\section{Testbeds}\label{sec:testbeds}

The nine methods of parent selection are applied on three testbeds. The \emph{6-D Rastrigin} (Section \ref{sec:testbeds_rastrigin}) and the \emph{12-DoF Arm Repertoire} (Section \ref{sec:testbeds_arm}) testbeds are two typical benchmarks for QD and evolutionary search more broadly. The maze generation task (Section \ref{sec:testbeds_mazes}) is closer to a real-world application for automated level design.

\subsection{6-D Rastrigin}\label{sec:testbeds_rastrigin}

Rastrigin is a classical benchmark for global optimization \cite{hansen2009bbob} that has often served as testbed for QD search \cite{justesen2019adaptive,flageat2020deep,fontaine2020covariance}. Rastrigin is a ``highly multimodal function with a comparatively regular structure for the placement of the optima'' \cite{hansen2009bbob}. This testbed is thus ideal for examining an algorithm's ability to detect global optima or to establish a good diversity overall. Following the practice of \cite{justesen2019adaptive,flageat2020deep}, we use the 6-dimensional version of the Rastrigin function.

\subsubsection{Experimental Setup}

The \textbf{genotype} is a vector of real-valued variables $\vec{x} = (x_1, x_2,\dots, x_6)$ with $\vec{x} \in [-5.12,5.12]^6$. 
A genotype's \textbf{fitness} is calculated via the 6-D Rastrigin function in Eq.~\eqref{eq:rastrigin}.
\begin{equation}\label{eq:rastrigin}
f(\vec{x}) = 60 + \sum_{i=1}^{6}(x_i^2 - 10 \cdot \cos(2 \pi x_i))
\end{equation}

Following \cite{justesen2019adaptive,flageat2020deep}, the \textbf{behavioral dimensions} for MAP-Elites are the genes $x_1$ and $x_2$. The feature map is subdivided into 100 equal segments along both dimensions, resulting in a grid of $10^4$ discrete cells with a side-length of $10.24 \cdot 10^{-2}$.
Mutation is applied uniformly to every gene by adding a random $r \in [-0.256,0.256]$ (5\% of the gene's value range), sampled from a uniform distribution. Mutated genes are truncated to the $[-5.12,5.12]$ value range.

\subsection{12-DoF Arm Repertoire}\label{sec:testbeds_arm}

The second testbed follows the Arm Repertoire robotic control task which has been proposed as a QD benchmark \cite{cully2018qualitydiversity}. The algorithm optimizes the angular positions of the different joints of a robotic arm, and the ``solution descriptor is defined as the final position of the gripper, which is then normalized according to a square bounding box to have values between 0 and 1'' \cite{cully2018qualitydiversity}. Based on preliminary experiments, we use 12 degrees of freedom for the Arm Repertoire task as the differences between some selection methods are less pronounced with few degrees of freedom.

\subsubsection{Experimental Setup}
The \textbf{genotype} is a vector of real-valued variables $\vec{\theta} = (\theta_1, \theta_2, \dots, \theta_{12})$ with $\vec{\theta} \in \begin{bmatrix}-\pi,\pi\end{bmatrix}^{12}$,
signifying that each joint can make a full $360^{\circ}$ rotation.
The goal is to equalize the joint angles and \textbf{fitness} is calculated by Eq.~\eqref{eq:arm_repertoire_fitness}, where $\mu$ is the mean angle of $\vec{\theta}$.
\begin{equation}\label{eq:arm_repertoire_fitness}
f(\vec{\theta}) = -\frac{1}{12} \sum_{i=1}^{12}(\theta_i - \mu)^2
\end{equation}

Following \cite{flageat2020deep}, the \textbf{behavioral dimensions} for MAP-Elites are the $x$ and $y$ coordinates of the final position of the robotic arm. These are calculated by Eq.~\eqref{eq:arm_repertoire_behavior}, where $l_i$ is the joint's length; $l = \frac{1}{12}$ for all joints, and thus the behavioral space is in the domain of $[-1,1]^2$. 
The feature map is subdivided into 100 equal parts along both dimensions, resulting in a grid of $10^4$ discrete cells with a side-length of $2 \cdot 10^{-2}$.
\begin{equation}\label{eq:arm_repertoire_behavior}
B(\vec{\theta}) =
\begin{bmatrix}
l_1 \cos(\theta_1) + l_2 \cos(\theta_1+\theta_2) + \cdots + l_{12}\cos(\sum_{i=1}^{12} \theta_i)\\
l_1 \sin(\theta_1) + l_2 \sin(\theta_1+\theta_2) + \cdots + l_{12}\sin(\sum_{i=1}^{12} \theta_i)
\end{bmatrix}
\end{equation}

Mutation is applied uniformly to every gene by adding a random $r \in [-0.1\pi,0.1\pi]$, sampled from a uniform distribution. All angles are wrapped to the $[-\pi,\pi]$ value range.

\subsection{Maze Generation}\label{sec:testbeds_mazes}

The final testbed is a simple level design task where QD could be beneficial for the designer \cite{gravina2019pcgqd}. The goal in this testbed is to generate perfect mazes in 2D orthogonal grids. Unlike previous testbeds, maze generation has no explicit functional or aesthetic dimensions. This testbed explores five different characteristics of mazes which a human designer could find interesting and tests all possible combinations of two feature dimensions and one fitness score for these characteristics. The maze generation testbed can give insights into the algorithms' performance, but also tests how the proposed QD variants apply to actual design problems.

\begin{figure}
\centering
\subfloat[Initial state \& solution]{
\label{fig:maze_mutation_1_2_initial_state_solution}
\includegraphics[width=0.14\textwidth]{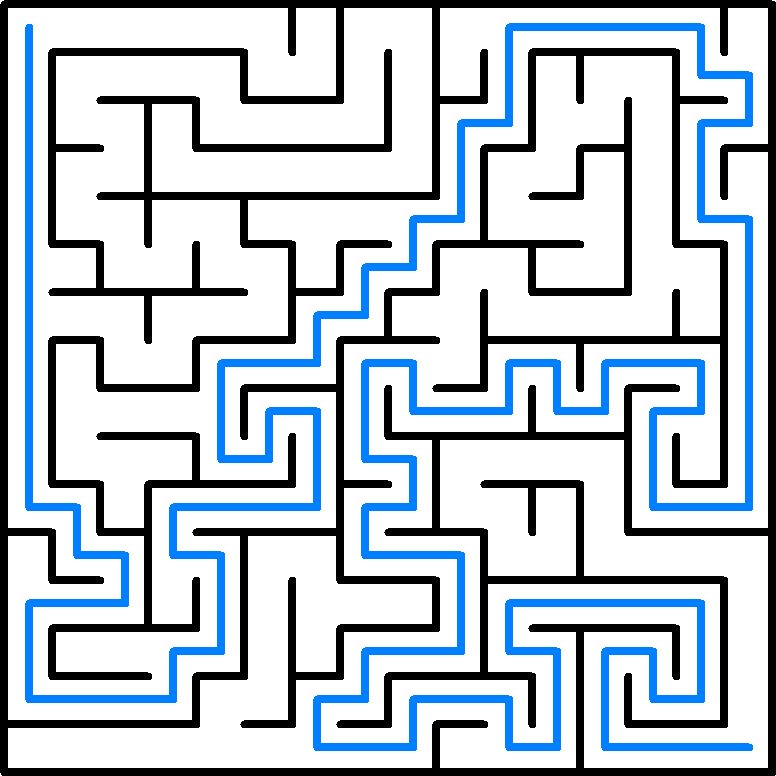}
}
\hfill
\subfloat[Mutation: destroyed cells]{
\label{fig:maze_mutation_2_1_destroyed_cells}
\includegraphics[width=0.14\textwidth]{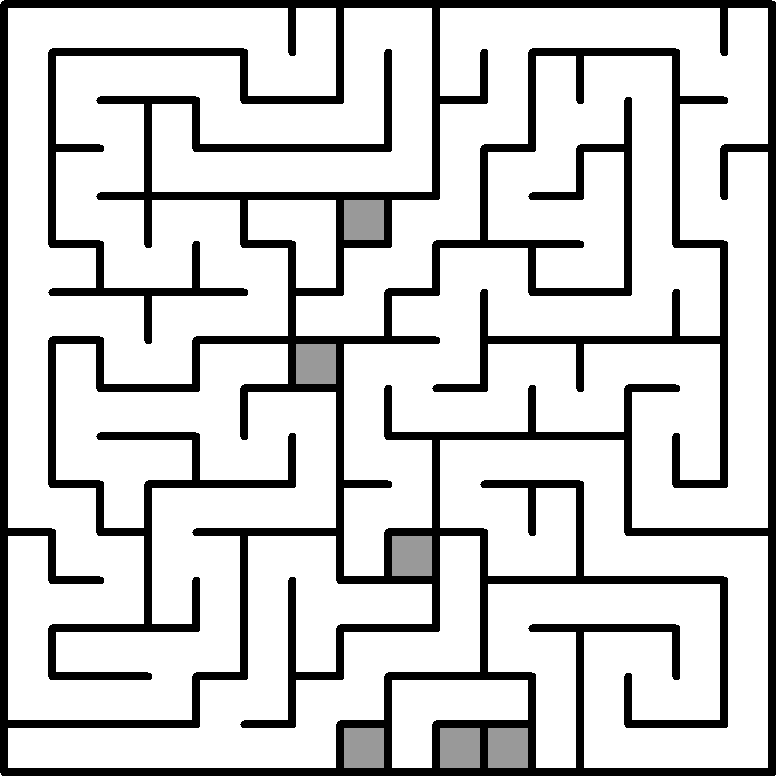}
}
\hfill
\subfloat[Repair step 1: reconnecting cells with RDFS]{
\label{fig:maze_mutation_3_1_reconnected_cells}
\includegraphics[width=0.14\textwidth]{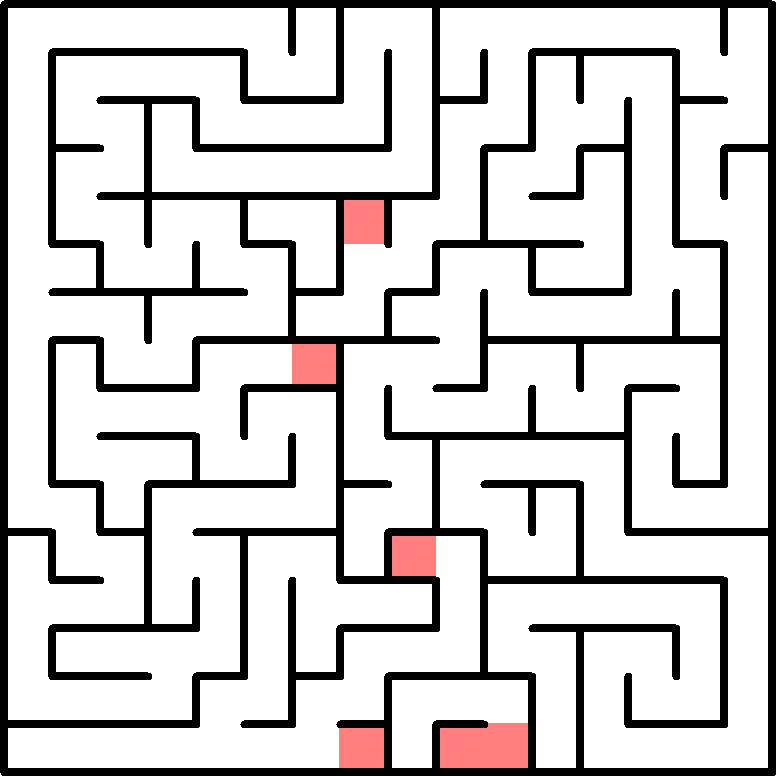}
}

\subfloat[Disconnected islands]{
\label{fig:maze_mutation_3_2_islands}
\includegraphics[width=0.14\textwidth]{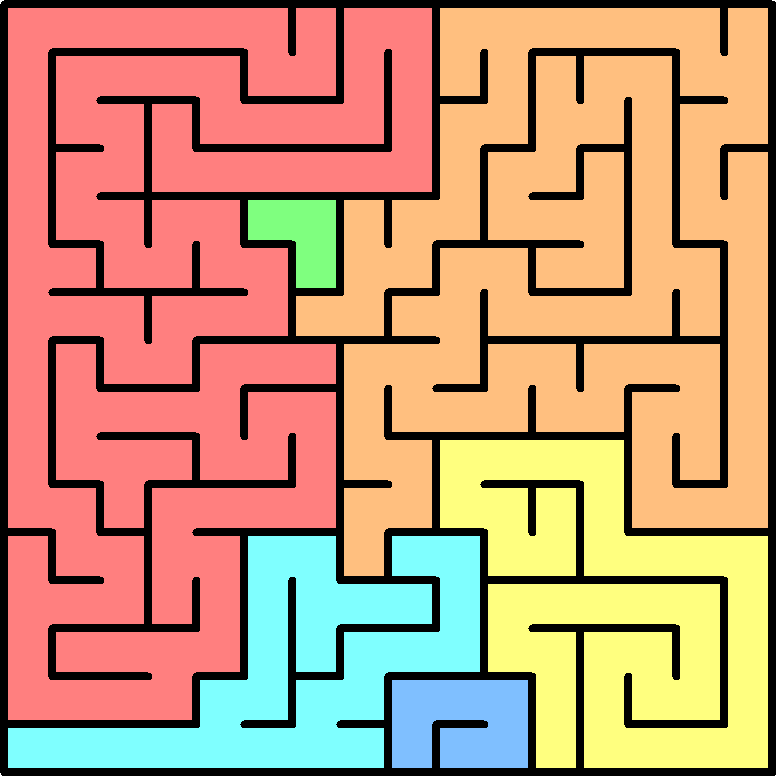}
}
\hfill
\subfloat[Repair step 2: reconnecting islands]{
\label{fig:maze_mutation_4_1_reconnected_islands}
\includegraphics[width=0.14\textwidth]{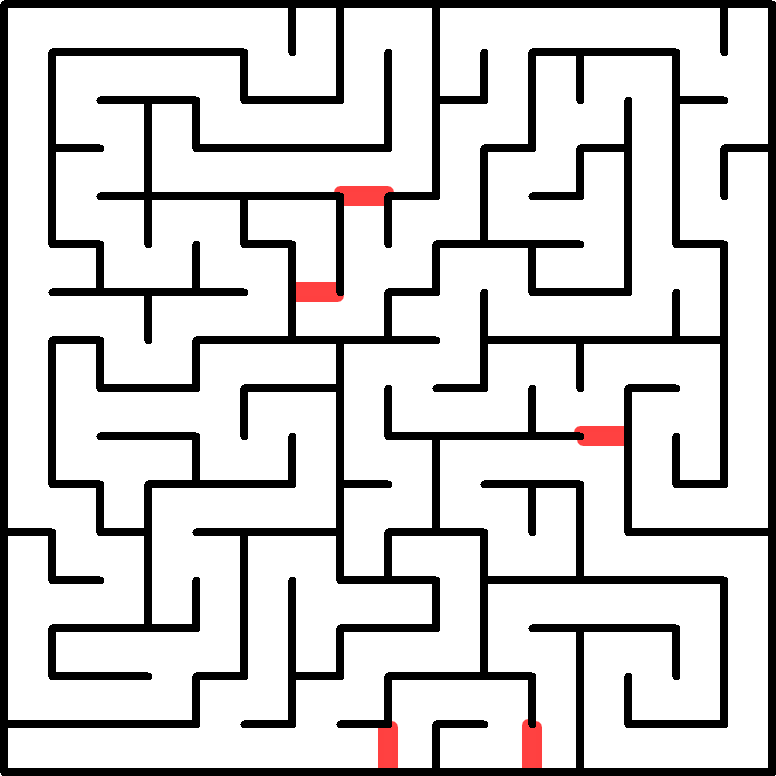}
}
\hfill
\subfloat[Final state \& solution]{
\label{fig:maze_mutation_4_2__final_state_solution}
\includegraphics[width=0.14\textwidth]{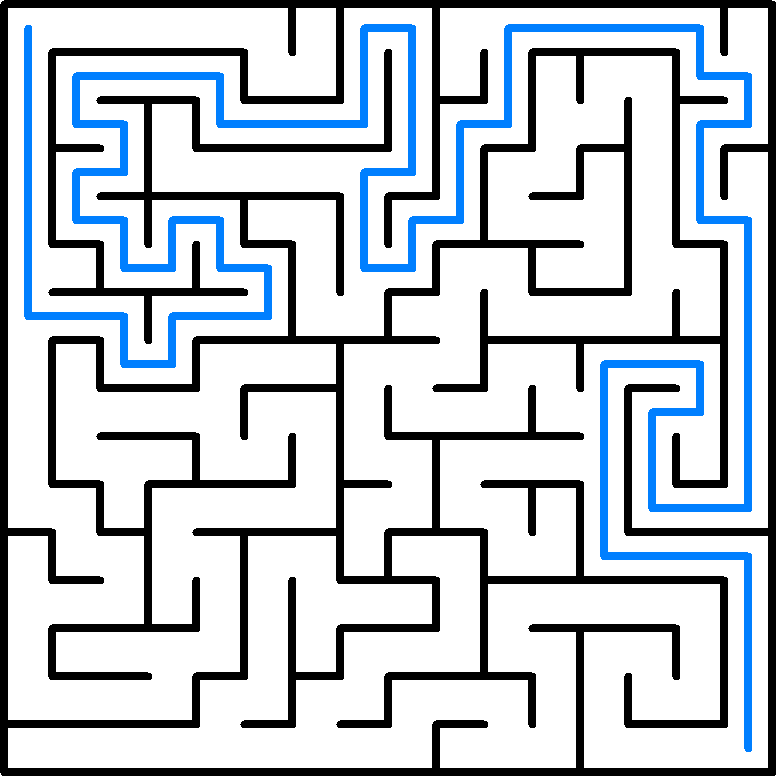}
}

\caption{Mutation and repair process for a maze of $16\times16$ tiles. Also shown is the shortest path from the top left corner to the bottom right corner used in the optimal path metric.} 
\label{fig:CEE_1___appendix}
\end{figure}

The maze is directly represented in the \textbf{genotype}, which is a 2-D matrix of integer IDs in the range of (0\dots15). Each ID determines the connectivity of that tile with all adjacent tiles in the four cardinal directions. The size of the genotype depends on the size of the maze, i.e. its width and height. A Random Depth-First-Search (RDFS) process, described in \cite{kozlova2015maze}, is used to generate the initial population and to repair individuals. The {mutation} operator destroys a number of tiles in the parent (setting every side of the tile to unconnected) and applies the RDFS process to repair the maze. Specifically, mutation iterates through every tile and has a 2\% chance of destroying it: ``destruction'' sets the tile's ID to one that is surrounded by walls, and all adjacent tiles change their ID accordingly so that their connection with the destroyed tile is removed. If no tile is mutated in this fashion, one random tile is chosen and destroyed as presented above. The resulting maze includes a number of disconnected cells and ``islands'' (see Fig.~\ref{fig:maze_mutation_2_1_destroyed_cells}); it must be repaired to become a perfect maze. The repair process first reconnects disconnected cells via RDFS (as shown in Fig.~\ref{fig:maze_mutation_3_1_reconnected_cells}), then merges remaining disconnected islands by randomly selecting an edge that connects all pairs of disconnected islands until all tiles are connected (as shown in Fig.~\ref{fig:maze_mutation_4_1_reconnected_islands}).

Based on our design sensibilities, a set of five metrics for mazes are formulated which are used interchangeably as quality or diversity measures. Two metrics assess the visual symmetry of the mazes. The \emph{horizontal symmetry} metric ($f_{H}$) reflects the maze along the $Y$ axis and calculates the number of tiles with the same connections between the original and the reflection. The \emph{bilateral symmetry} metric ($f_{B}$) measures similarity of the maze with both reflections on the $X$ axis and the $Y$ axis. Two metrics assess patterns of tiles with two connections (corridors) which are usually common in such mazes: the \emph{corner} metric ($f_{L}$) counts the number of corridors where the two connections are at a right angle, and the \emph{straight} metric ($f_{I}$) counts the number of corridors where the two connections are at a straight angle. All of the above four metrics are normalized to the total number of tiles of the maze. The final metric is for \emph{optimal path} ($f_{P}$) and assumes that the maze's start is the top-left corner and its end is the bottom-right corner (see Fig.~\ref{fig:maze_mutation_1_2_initial_state_solution}), and that the shortest path between the two corners should ideally cover half of the maze's tiles. This ad-hoc assumption allows the maze's functional aspects to be evaluated rather than only its visual properties. The $f_{P}$ metric is calculated as $1-\left\lvert{2P}/T - 1\right\rvert$ where $P$ is the shortest path between the top-left and bottom-right corners; $T$ is the number of tiles of the maze as the product of its height and width.

\subsection{Experimental Protocol}

For the Rastrigin and Arm Repertoire experiments we perform 100 independent evolutionary runs. A population of 100 random individuals is used to populate the initial feature map. In each iteration an individual is selected according to the current method (see Section \ref{sec:methodology}) and produces one offspring which replaces an existing elite, occupies a previously empty cell, or is discarded. 
The resolution of the feature map for these two experiments is $100\times100$ cells. Evolutionary runs finish after $10^6$ evaluations (i.e. iterations). 

For the Maze Generation experiments, all possible combinations of metrics are tested as either a fitness function or two aesthetic measures. This results in 30 possible combinations, each of which is tested in 100 independent evolutionary runs which finish after $10^5$ evaluations for the sake of brevity. The resolution of the feature map for these maze generation experiments is $50\times50$ cells. Otherwise, the evolutionary process is identical to the one described above.

The analysis of the results is performed across five performance metrics typical for the assessment of QD methods. The analysis implements the metrics of \textbf{Global Performance}, \textbf{Global Reliability}, \textbf{Precision} and \textbf{Coverage} which are described in \cite{Mouret2015IlluminatingSS}, as well as the \textbf{QD-score} described in \cite{pugh2016quality} and calculated as the fitness sum of all the currently populated cells.
In addition, a metric is introduced not to assess performance but rather to observe how different biases influence which cells are selected: \textbf{Selection Entropy} expresses the degree of uniformity of selections on the map. It is based on Shannon's entropy applied on the selection map, calculated as  $-\frac{1}{\log{N_c}} \sum_{i=1}^{N_c}\left(\frac{n_c(i)}{N_s} \log \left(\frac{n_c(i)}{N_s}\right)\right)$ where $N_c$ is the total number of cells, $n_{c}(i)$ is the times that any individual of cell $i$ was selected, and $N_s$ is the total number of selections.

Given that we test nine selection methods across five performance metrics, reporting all the results becomes cumbersome. We focus instead on the number of methods that a selection method is significantly better (higher) than in a performance metric. All tests are performed on Welch's $t$-test at $\alpha=0.05$ significance threshold. Since a selection method is compared against 8 other methods, we apply the Bonferroni correction \cite{dunnett1955multiple} to test for significance. To assess overall performance across evaluations, we measure the area under the curve (AUC) from the start of evolution until the end of the run---i.e. after $10^6$ evaluations for Rastrigin and Arm Repertoire, and after $10^5$ for Maze Generation.

\section{Results}\label{sec:results}

This section highlights how the proposed selection methods influence QD search for each of the three testbeds presented in Section \ref{sec:testbeds}. Results are averaged across 100 runs. We test for significance by employing Welch's $t$-tests at a significance threshold $\alpha=0.05$, applying the Bonferroni correction for multiple comparisons.

The source code for all experiments, which can be used to reproduce or expand on the results, can be found at \url{https://github.com/konsfik/Monte-Carlo-Elites} .

\subsection{6-D Rastrigin}\label{sec:results_rastrigin}

\begin{figure}
\centering
\includegraphics[width=\columnwidth]{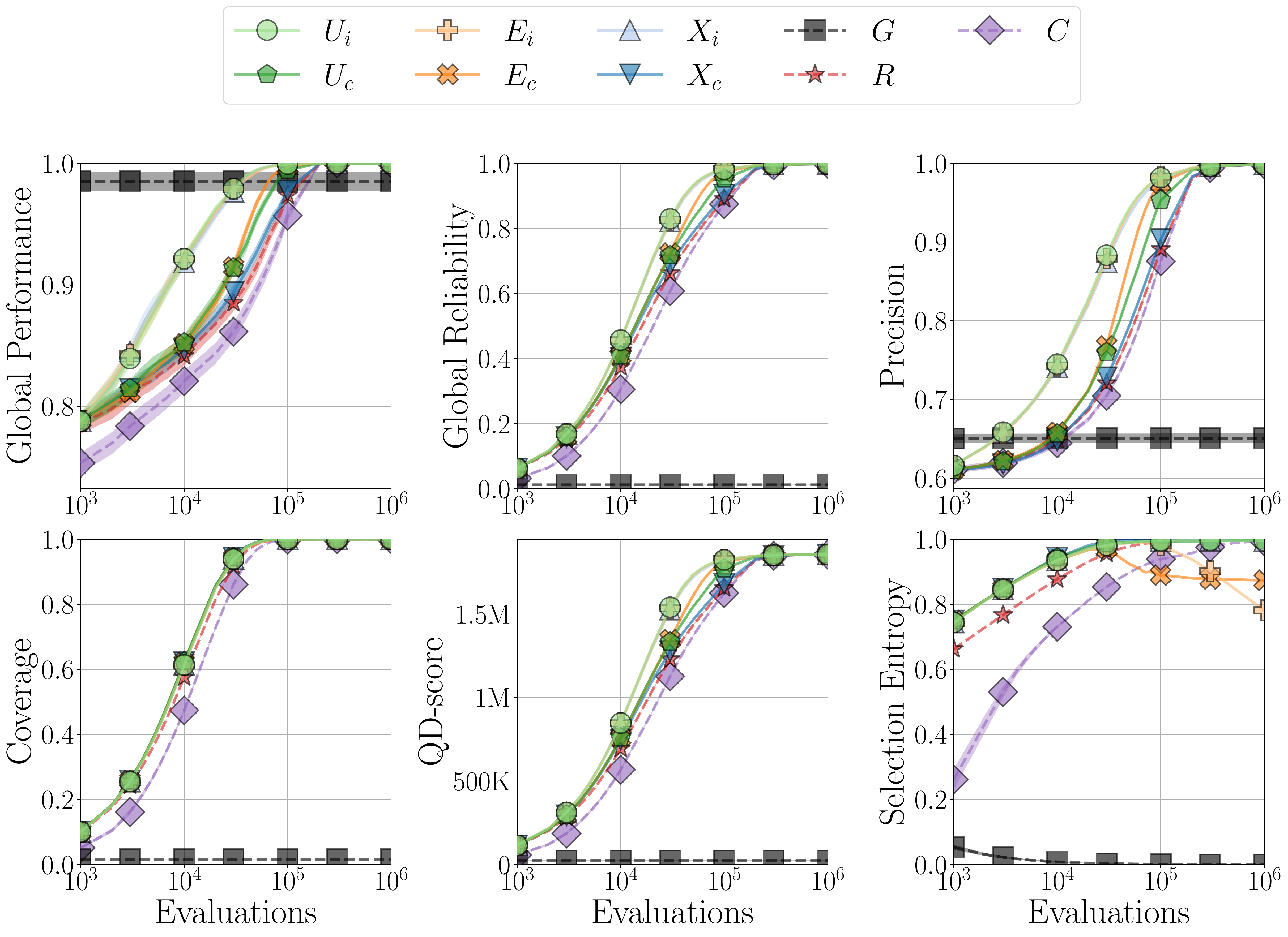}
\caption{Rastrigin 6D: progression of performance metrics between $10^3$ and $10^6$ evaluations. Results are averaged from 100 runs; shaded areas show the 95\% confidence interval.}
\label{fig:rastrigin_charts}
\end{figure}

Figure \ref{fig:rastrigin_charts} shows how each performance metric fluctuates between $10^3$ and $10^6$ evaluations, visualized in logarithmic scale. As expected, the greedy fitness-based selection performs well in terms of global performance early on but converges to local optima and is eventually outperformed by all other methods after $10^5$ and $10^6$ evaluations. Interestingly, all cell-based selection methods ($E_c$, $X_c$, $U_c$) outperform other methods (but not each other) in terms of coverage in early stages of evolution but eventually all methods (except $G$) reach 100\% coverage. Interestingly, the curiosity baseline is much slower in covering the entire feature map but also in finding a global optimum (global performance). Global reliability, precision and QD-score all share similar patterns, with individual-based selection methods ($E_i$, $X_i$, $U_i$) outperforming other methods from $10^4$ evaluations and after. 
In terms of global reliability, precision and QD-score, $E_i$ performs significantly better than all other methods after $10^5$ evaluations, but after $10^6$ evaluations it is $U_i$ that outperforms all other methods.

\begin{table}[t]
\centering
\caption{\textbf{Rastrigin 6D:} The number under each column indicates the times a selection method yields significantly higher AUC values in the row's metric, compared to the AUC values of the remaining 8 selection methods.
The best methods per metric appear in bold.}
\label{tab:auc_rastrigin}
\begin{tabular}{ l | c | c | c | c | c | c | c | c | c }
Method & $U_i$ & $U_c$ & $E_i$ & $E_c$ & $X_i$ & $X_c$ & $G$ & $R$ & $C$ \\ \hline \hline
Glob. Perf. & \textbf{6} & 3 & \textbf{6} & 4 & \textbf{6} & 1 & 0 & 1 & 0 \\ \hline
Glob. Rel.  & \textbf{7} & 4 & \textbf{7} & 5 & 6 & 3 & 0 & 2 & 1 \\ \hline
Precision   & \textbf{7} & 4 & \textbf{7} & 5 & 6 & 3 & 0 & 2 & 1 \\ \hline
Coverage    & 3 & \textbf{7} & 3 & 6 & 3 & \textbf{7} & 0 & 2 & 1 \\ \hline
QD-score    & \textbf{7} & 4 & \textbf{7} & 5 & 6 & 3 & 0 & 2 & 1 \\

\end{tabular}
\end{table}

Table \ref{tab:auc_rastrigin} shows how each method compares in terms of the AUC of each performance metric until $10^6$ evaluations. We observe that both $E_i$ and $U_i$ perform well in all performance metrics except coverage, outperforming all other (7) methods in terms of global reliability, precision and QD-score. As evidenced from Fig.~\ref{fig:rastrigin_charts}, the curiosity baseline is outperformed by all other methods in terms of global performance. On the other hand, both $U_c$ and $X_c$ outperform all other (7) methods in terms of coverage. It is evident that for the deceptive, multimodal fitness landscape of Rastrigin the individual-based selection methods are best at finding a large number of good solutions; the cell-based selection methods, however, are better at exploring the landscape.

We observe in Fig.~\ref{fig:rastrigin_charts} that the selection entropy is higher for all methods proposed in this paper than the random selection until $10^4$ evaluations, meaning that all cells over the course of evolution were selected almost equally often. Most methods (including the $R$ baseline) reach nigh-uniform selection after $10^5$ evaluations, unsurprisingly when the coverage metric reaches 100\%. Interestingly, however, when coverage reaches 100\% the exploitation-only approaches $E_i$ and $E_c$ start focusing on specific areas of the search space and show a drop in their selection entropy score. It is even more interesting that the drops follow a different trend: for $E_c$ selection entropy drops early but stabilizes, while for $E_i$ selection entropy drops later but continues dropping. This is likely because for $E_i$ new individuals in previously occupied cells receive absolute priority (as $n_i=0$ in Eq.~\ref{eq:selection_ucb}), while for $E_c$ there is no distinction. When coverage reaches 100\%, $E_i$ continues focusing on new individuals while the selection strategy for $E_c$ reaches equilibrium. Finally, it is worth noting that the curiosity baseline has a very different selection entropy than conceptually similar methods (e.g.~$E_i$) and its selection entropy increases far slower than other methods---likely due to the lower coverage in early stages of evolution for curiosity.

\begin{figure}
\centering
\includegraphics[width=0.85\columnwidth]{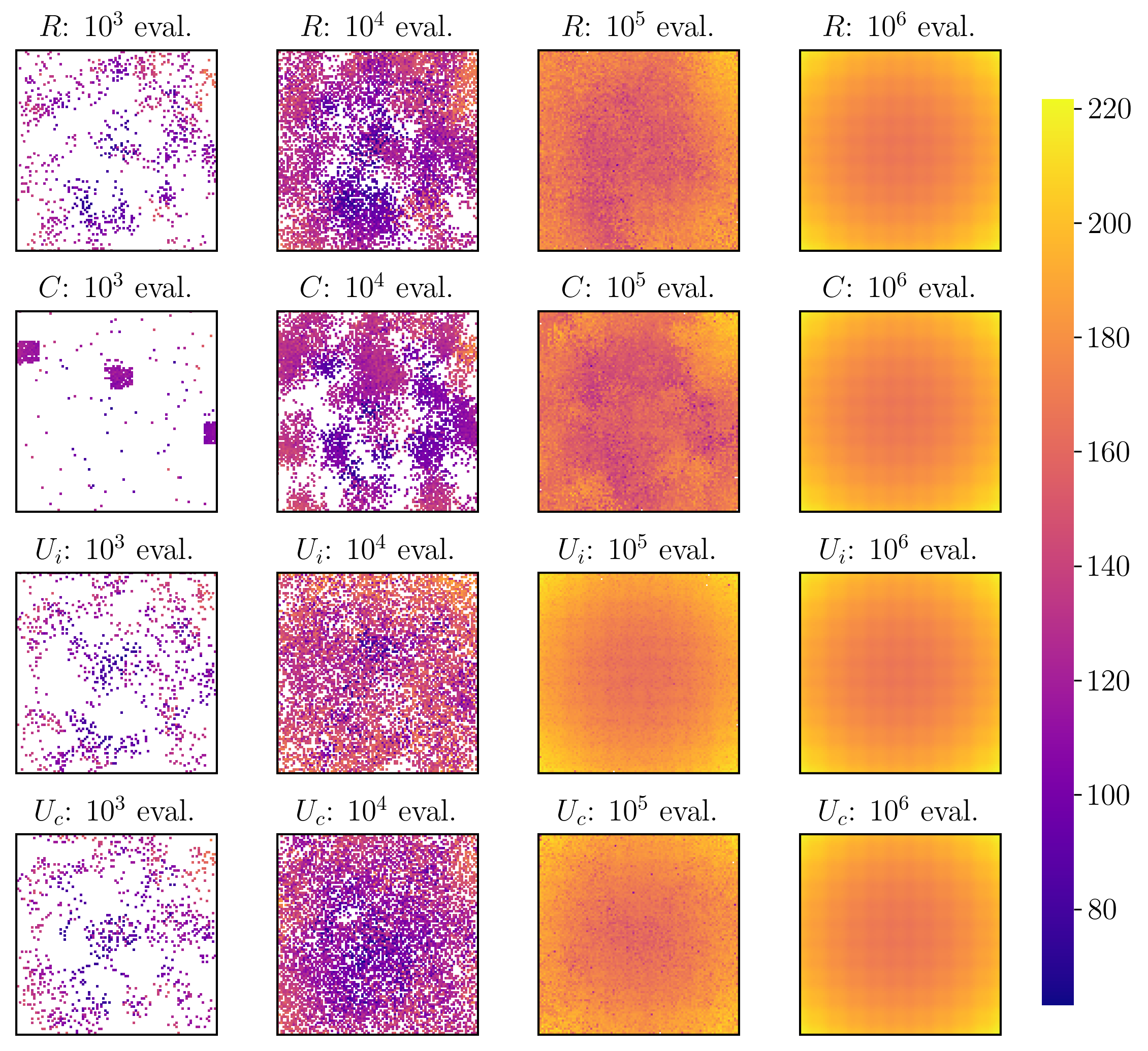}
\caption{Rastrigin 6D: Fitness and selection heat-maps for a single run, captured after $10^{3}$, $10^{4}$, $10^{5}$, $10^{6}$ evaluations.}
\label{fig:rastrigin_heatmaps_single_run}
\end{figure}

Fig.~\ref{fig:rastrigin_heatmaps_single_run} shows the feature maps of the two UCB variants and the uniform and curiosity baselines across evaluation thresholds in a sample run (the first run of the experiment). The higher coverage of $U_i$ and especially $U_c$ is visible after $10^3$ and $10^4$ evaluations. The feature map for $C$ at $10^3$ generations is especially interesting, as only a few clusters within the space are explored, and selection has exclusively focused on those early parents at the center of those clusters. Moreover, the areas of low fitness are far more pronounced in the $R$ and $C$ baselines after $10^5$ evaluations. 

\subsection{12-DoF Arm Repertoire}\label{sec:results_arm}

\begin{figure}
\centering
\includegraphics[width=\columnwidth]{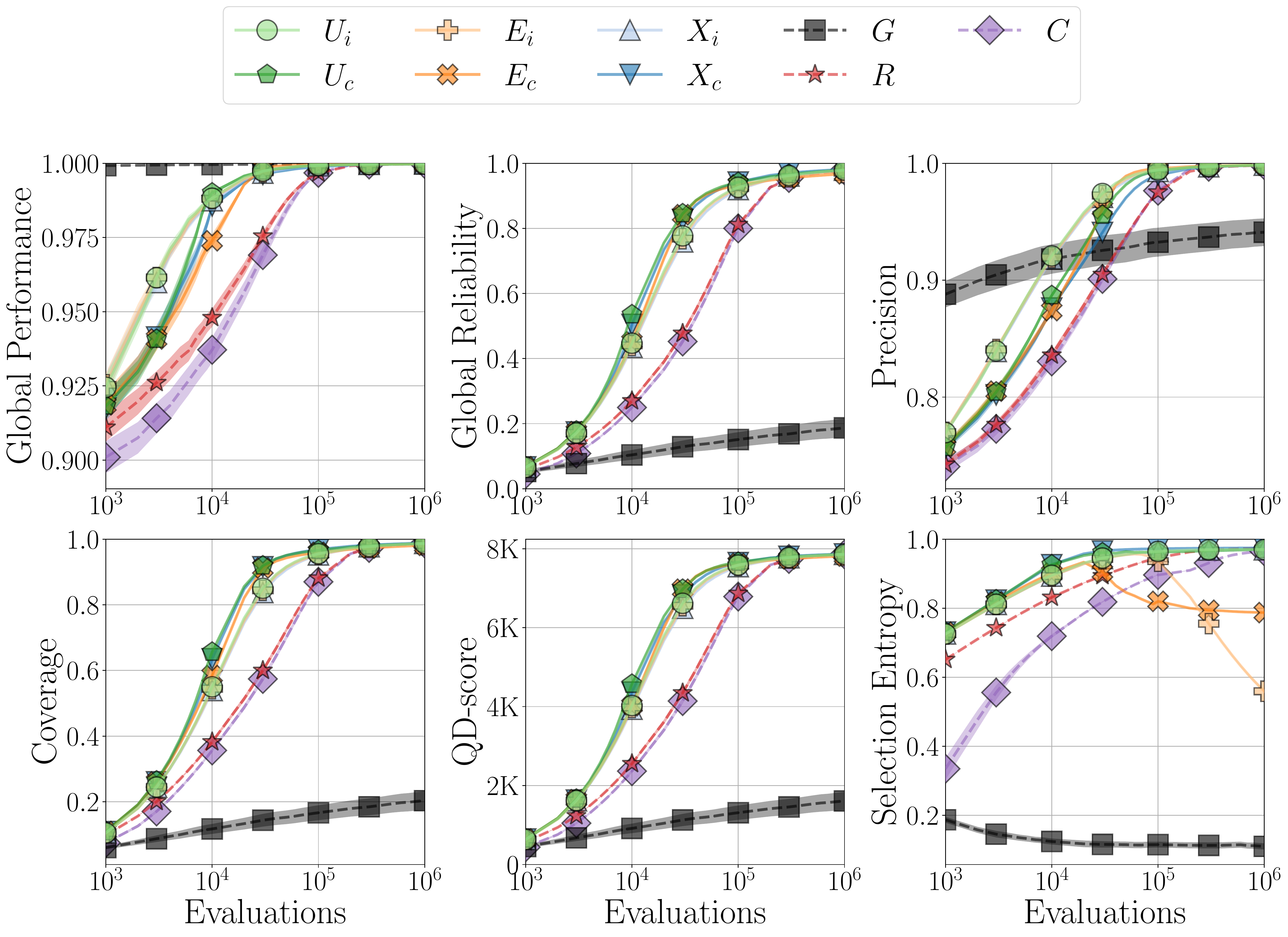}
\caption{12-DoF Arm Repertoire: progression of performance metrics between $10^3$ and $10^6$ evaluations. Results are averaged from 100 runs; shaded areas show the 95\% confidence interval.}
\label{fig:arm_charts}
\end{figure}

\begin{table}[t]
\centering
\caption{{12-DoF Arm Repertoire:} The number under each column indicates the times a selection method yields significantly higher AUC values in the row's metric, compared to the AUC values of the remaining 8 selection methods. The best methods per metric appear in bold.}
\label{tab:auc_arm}
\begin{tabular}{
 l | c | c | c | c | c | c | c | c | c 
}
Method & $U_i$ & $U_c$ & $E_i$ & $E_c$ & $X_i$ & $X_c$ & $G$ & $R$ & $C$ \\ \hline\hline
Glob. Perf. & 6 & 4 & 7 & 3 & 4 & 2 & \textbf{8} & 1 & 0 \\ \hline
Glob. Rel.  & 6 & 7 & 3 & 3 & 5 & \textbf{8} & 0 & 1 & 1 \\ \hline
Precision   & 7 & 4 & \textbf{8} & 5 & 6 & 3 & 0 & 1 & 1 \\ \hline
Coverage    & 6 & 7 & 3 & 4 & 4 & \textbf{8} & 0 & 2 & 1 \\ \hline
QD-score    & 6 & \textbf{7} & 3 & 4 & 4 & \textbf{7} & 0 & 1 & 1 \\
\end{tabular}
\end{table}

Figure \ref{fig:arm_charts} shows how each performance fluctuates between $10^3$ and $10^6$ evaluations for the Arm Repertoire testbed. As this testbed does not have the deceptive fitness landscape of Rastrigin, the greedy baseline consistently outperforms all other methods in terms of global performance. Besides an early lead in terms of precision, however, this greedy approach is outperformed in all other metrics. Unlike the Rastrigin testbed, cell-based approaches seem to perform better in most metrics. In terms of global reliability, coverage, and QD-score, $U_c$ outperforms all other methods at $10^4$ evaluations, but the lead changes at $10^6$ evaluations as $X_c$ outperforms all other methods. In terms of precision, exploitation-based approaches ($E_i$, $E_c$) and $U_i$ perform significantly better than the other methods from $10^5$ evaluations and after.

Table \ref{tab:auc_arm} shows how each method compares in terms of the AUC of each performance metric until $10^6$ evaluations. As expected, the greedy approach outperforms all other methods in terms of global performance. All selection methods introduced in this paper perform better than uniform ($R$) and curiosity ($C$) baselines in terms of global performance, although individual-based approaches perform better in that metric. Interestingly, $X_c$ outperforms all other methods in terms of global reliability and coverage, while in terms of QD-score only $U_c$ performs comparably to $X_c$. $U_i$ is the only method to reach fairly high scores in all metrics, although it is always outperformed by another method in each metric.

\begin{figure}
\centering
\includegraphics[width=0.85\columnwidth]{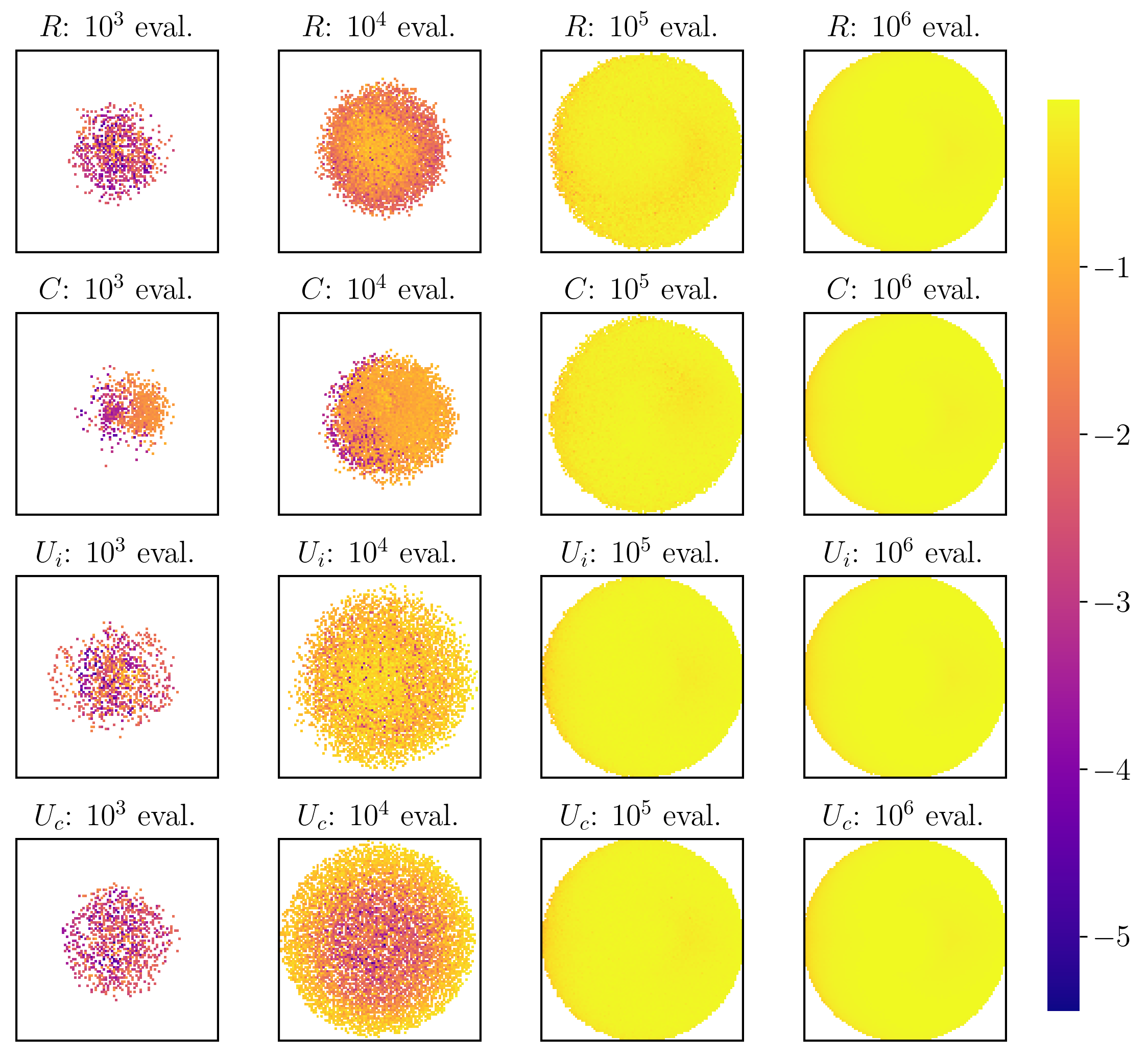}
\caption{12-DoF Arm Repertoire: Fitness heat-maps for one run captured after $10^{3}$, $10^{4}$, $10^{5}$, $10^{6}$ evaluations.}
\label{fig:arm_repertoire_heatmaps_single_run}
\end{figure}

It is worth noting that the selection entropy in this testbed (Fig.~\ref{fig:arm_charts}) has a very similar trend as for Rastrigin, especially the drop in selection entropy for $E_c$ and---later and more abruptly---for $E_i$ when coverage plateaus. The curiosity baseline follows similar patterns as in Rastrigin and its selection entropy is low until later stages of evolution. Figure~\ref{fig:arm_repertoire_heatmaps_single_run} shows the feature maps of a sample run for Arm Repertoire. It is immediately obvious that at $10^4$ evaluations the $R$ baseline has found highly fit solutions in the center of the search space, while $U_c$ has found highly fit solutions at the edges of the discovered search space but solutions at the center are less fit. Since $U_c$ tends to select newly discovered cells more often, it is not surprising that it discovers fit individuals around the edges of the ever-expanding feature map. The $C$ baseline at $10^4$ evaluations has a similar coverage to $R$, but seems to have discovered fit individuals mainly on the right half of the feature map, likely because it always selects the same few parents (already visible at $10^3$ evaluations). The $U_i$ selection method after $10^4$ evaluations is the most homogeneous, while eventually all approaches cover the entire circle that the arm can move in and find highly fit solutions in that space. 

\subsection{Maze Generation}\label{sec:results_maze}

In the maze generation testbed, we test a broad variety of quality and diversity dimensions, as well as multiple sizes of the genotype and phenotype. All results reported in this section are aggregated across 30 combinations of different feature dimensions and fitness functions (using the five metrics of Section \ref{sec:testbeds_mazes}) in two different maze lattices ($8{\times}8$ tiles and $16{\times}16$ tiles). Since a total of 60 treatments (of 100 evolutionary runs each) are being compared, the results can be more conclusive for the behavior of the selection methods.

\begin{figure}
\centering
\includegraphics[width=\columnwidth]{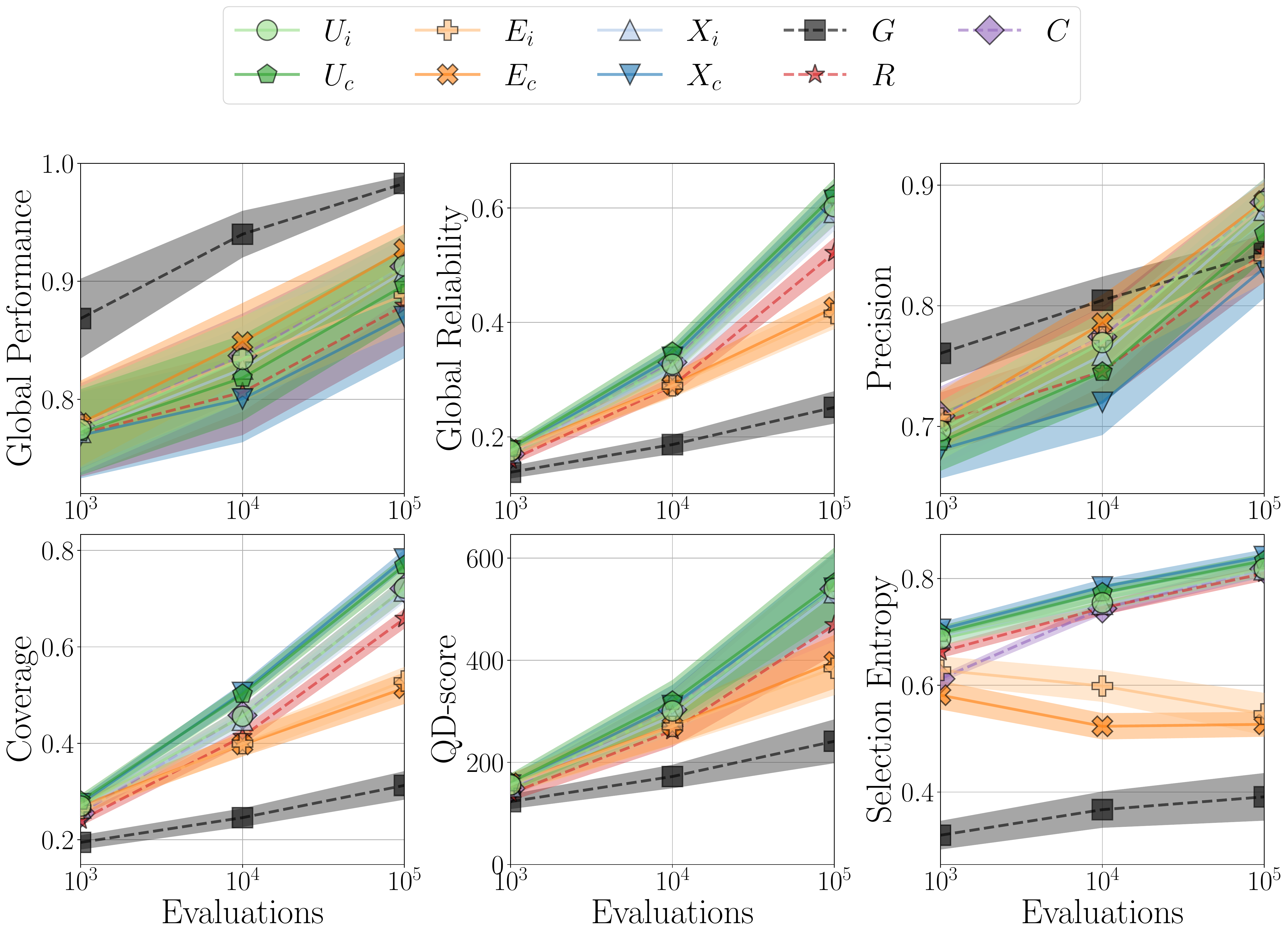}
\caption{Maze generation: progression of performance metrics between $10^3$ and $10^5$ evaluations. Metrics are averaged from 60 treatments, and shaded areas show the 95\% confidence intervals of the average scores per treatment.}
\label{fig:mazes_progress}
\end{figure}

Figure \ref{fig:mazes_progress} shows how each performance fluctuates between $10^3$ and $10^5$ evaluations, averaged from all 60 treatments. It is evident that both the best fitness and the QD score have not plateaued, and more generations could shed more light on performance. Unlike in the other two testbeds, we observe that coverage for the curiosity baseline in this case is better than the uniform baseline, while its selection entropy is comparable to that of $R$. Moreover, selection entropy for $E_i$, $E_c$ is surprisingly low throughout evolution.

Table \ref{tab:auc_maze} shows how each method compares in terms of the AUC of each performance metric until $10^5$ evaluations. In terms of global performance, the greedy fitness-based approach usually outperforms the other methods while interestingly $E_c$ outperforms $C$ in 28 treatments and is outperformed in 8. The curiosity baseline in this testbed performs quite well, and outperforms most methods including $E_i$ (which is conceptually similar) and $E_c$ in terms of global reliability, coverage and QD score. However, in terms of QD score the cell-based UCB approach ($U_c$) is superior, outperforming $C$ in 42 treatments and $X_c$ in 36 treatments. In terms of coverage, the exploration-based $X_c$ outperforms all methods in 58 or more treatments---except $U_c$, which it only outperforms in 34 treatments.

As a general takeaway from this analysis, exploitation-only methods seem to perform worse than the UCB or exploration-only methods, with $E_c$ performing well in terms of global performance and precision. Curiosity in this testbed performs comparably to $U_i$ but both methods are outperformed by others in every metric. Interestingly, cell-based approaches tend to explore the search space faster and find fairly fit individuals in all cells (as evidenced by high QD-scores); the focus on exploration, however, does not allow them to evolve highly performing individuals (especially $X_c$).

\begin{table}[t]
\centering
\caption{\textbf{Maze Generation:} The number under each column indicates the times a selection method yields significantly higher AUC values in the row's metric, compared to the AUC values of the remaining 8 selection methods for a total of 480 comparisons across setups.
The best method per metric is shown in bold.
}
\label{tab:auc_maze}
\begin{tabular}{
 l |c@{ }|c@{ }|c@{ }|c@{ }|c@{ }| c@{ }|c@{ }|c@{ }|c@{ }
}
Method & $U_i$ & $U_c$ & $E_i$ & $E_c$ & $X_i$ & $X_c$ & $G$ & $R$ & $C$  \\ \hline\hline
Glob. Perf. & 219 & 106 & 145 & 276 & 158 & 8   & \textbf{384} & 39  & 233 \\ \hline
Glob. Rel.  & 306 & \textbf{425} & 82  & 95  & 241 & 364 & 2   & 158 & 309 \\ \hline
Precision   & 325 & 128 & 172 & \textbf{359} & 227 & 19  & 286 & 97  & 355 \\ \hline
Coverage    & 283 & 426 & 84  & 82  & 240 & \textbf{451} & 1   & 169 & 272 \\ \hline
QD-score    & 318 & \textbf{420} & 81  & 102 & 249 & 344 & 2   & 155 & 321 \\
\end{tabular}
\end{table}

\section{Discussion}\label{sec:discussion}

Results in Section \ref{sec:results} have shown that overall any of the selection methods proposed in this paper outperforms the uniform selection in the original MAP-Elites across all experiments. Moreover, in Rastrigin and Arm Repertoire the curiosity baseline is slower to discover new cells than other methods, although its coverage is better than exploitation-only methods in the maze generation testbed. However, coverage in all three testbeds was higher for parent selection methods that assess UCB or exploration based on the cells occupied by each individual, rather than the individual itself. We hypothesize that this is due to the fact that newly filled cells always have priority. The ranking-based (versus random or roulette-wheel) selection applied in all proposed methods in this paper forces evolution to explore the newly discovered areas of the search space, until newer areas are discovered or until repeated selections of such individuals fail to yield good offspring. When selected parents are newly discovered cells (e.g. via cell-based selection) at the edge of the discovered search space, they are more likely to discover new cells, which leads to a positive feedback loop. This by-product of the UCB formula is likely why coverage is so superior early on, as seen in feature maps in all experiments. 

Comparing between methods, moreover, it is not surprising that exploration-only methods lead to higher coverage while exploitation-only methods lead to higher fitness in the fewer individuals discovered (precision). The UCB score, which combines exploration and exploitation, seems to lead to the best balance and the highest QD-scores, i.e. a good and diverse archive of elites. The only odd finding is that choosing parents based on the cell they are in or based on the individual itself has a strong impact that leads to one outperforming the other depending on the problem. In the deceptive landscape of Rastrigin, individual-based approaches work better; in all other experiments (including 60 variants of maze generation tasks) the cell-based approach leads to better and more diverse individuals but a lower maximum fitness. It is worth noting that in Rastrigin we use part of the genotype as feature dimensions, while in the other two testbeds the feature dimensions are only indirectly influenced (in a non-linear fashion) by the genotype. Based on selection heatmaps examined, the individual-based approach forces evolution to select parents in areas of the search space that were explored before, when a new individual is inserted there (even if there were many individuals there before it). This apparently can lead to breakthroughs in problems with a deceptive landscape (such as Rastrigin). In spaces that are ``easy'' to navigate, however, the way that cell-based selection focuses on new or recently filled cells (which are often on the edges of the feature map) leads to higher coverage and thus quick optimization of many elites.

While the focus was on the application of UCB for parent selection, this paper has also made contributions on the use of offspring survival as a reward mechanism. As noted in Section \ref{sec:background}, offspring survival has been identified as an important measure \cite{cully2018qualitydiversity,gaier2020discovering} and as a milestone in an individual's lifetime \cite{ecoffet2021goexplore}. This paper contributes to earlier work by exploring how survival is assessed (based on the individual or the cell). Comparisons of the proposed rank-based selection methods with the stochastic selection via an individual-based curiosity score \cite{cully2018qualitydiversity} show that there are important differences in terms of performance, coverage, and general behavior (as shown in the feature maps). We hypothesize that an important factor for these differences is the fact that new cells or individuals receive absolute priority for selection via Eq.~\eqref{eq:selection_ucb}-\eqref{eq:selection_exploration}; especially in cell-based approaches, this priority for new cells leads to a much faster exploration of the feature space.
We also conducted preliminary experiments with variants of offspring survival, such as the times an offspring replaced an elite, or the times an offspring discovered a new cell; these experiments yielded performance very close to the current metric which essentially combines the two. In other experiments that used fitness directly as the $w$ of Eq.~\eqref{eq:selection_ucb}, issues arose as the fitness in different testbeds had different value ranges and ad-hoc weights for $\lambda$ or normalization processes would be needed to balance the exploitation and the exploration components. It is also worth noting that while the selection mechanism introduced lacks the archive of past individuals of MENOV \cite{pugh2016quality}, historical trends are considered in the way the survival rate and selection bias is computed across all generations (especially for cell-based methods). That said, it is likely that there are other ways to assess both exploitation and exploration in Eq.~\ref{eq:selection_ucb} which can lead to better behaviors and can be examined in future work.

\section{Conclusion}\label{sec:conclusion}

This paper framed parent selection in MAP-Elites as a problem of exploration-exploitation balance, and examined the impact of selection via upper confidence bound on QD search across three different testbeds. Moreover, the exploitation component was assessed not directly on quality characteristics (e.g. based on the fitness score) but indirectly, based on the survival chances of the offspring of the individual. Several hypotheses are tested in the experiments, importantly whether the elite to choose should be based on its own offsprings' survival rate or based on all elites that ever existed in this part of the search space. Results indicate that a higher coverage can be attained by prioritizing exploration, while a generally better archive of diverse and good individuals is collected by maintaining a balance between exploration and exploitation. Regarding cell-based versus individual-based approaches, it seems that cell-based approaches are more robust and perform better in more experiments, especially in regards to covering the search space. Importantly, all methods that prioritize selection of newly inserted individuals or individuals who have not been selected often so far lead to improved performance across all metrics compared to the ``vanilla'' MAP-Elites that performs uniform selection.

\begin{acks}
This project has received funding from the EU’s Horizon 2020 programme under grant agreement No 951911.
\end{acks}

\bibliographystyle{ACM-Reference-Format}
\bibliography{mapelites,liapis}

\end{document}